\documentclass[journal]{IEEEtai}

\usepackage[colorlinks,urlcolor=blue,linkcolor=blue,citecolor=blue]{hyperref}

\usepackage{color,array}

\usepackage{graphicx}

\usepackage{soul}
\usepackage{lettrine}
\usepackage{cite}
\usepackage{supertabular,booktabs}
\usepackage{float}
\usepackage{comment}
\usepackage{makecell}
\usepackage{graphicx}
\usepackage{array}
\usepackage{amsfonts}
\usepackage{pdfpages}
\usepackage{amsmath}
\usepackage{algorithm}
\usepackage{algpseudocode}
\usepackage{xcolor}

\usepackage{subfig}
\usepackage{amssymb}
\usepackage{relsize}

\begin{document}

\title{SCRUB-FL: Sanitizing and Cleansing Representations via Unlearning of Backdoors}

\author{
    Osama Wehbi, Sarhad Arisdakessian, Omar Abdel Wahab, Azzam Mourad, Hadi Otrok\\
    \thanks{
        Osama Wehbi is with the Department of Computer and Software Engineering, Polytechnique Montréal, Montreal, Quebec, Canada (e-mail: osama.wehbi@etud.polymtl.ca).

        Sarhad Arisdakessian is with the Department of Computer and Software Engineering, Polytechnique Montréal, Montreal, Quebec, Canada (e-mails: sarhad.arisdakessian@etud.polymtl.ca).

        Omar Abdel Wahab is with the Department of Computer and Software Engineering, Polytechnique Montréal, Montreal, Quebec, Canada (e-mails: omar.abdul-wahab@polymtl.ca).

        Azzam Mourad is with the Department of Computer Science, Khalifa University, Abu Dhabi, UAE (e-mails: azzam.mourad@ku.ac.ae).
        
        Hadi Otrok is with the Department of Computer Science, Khalifa University, Abu Dhabi, UAE (e-mails: hadi.otrok@ku.ac.ae).
          
    }
}


\maketitle

\begin{abstract}
Federated Learning (FL) enables collaborative model training without sharing raw data, making it a promising paradigm for privacy-sensitive applications. However, its decentralized nature makes it inherently vulnerable to backdoor attacks, where malicious clients embed hidden triggers into local training data to manipulate model predictions. Existing defenses mainly operate during before and during aggregation cannot fully eliminate backdoor behaviors that persist in the converged global model. Moreover, the effectiveness of post-training sanitization is often limited by the server’s lack of knowledge of trigger patterns or poisoned clients after convergence, resulting in residual backdoor behaviors or accuracy degradation due to neuron entanglement. To address this limitation, we propose SCRUB-FL (Sanitizing and Cleansing Representations via Unlearning of Backdoors), a two-phase solution for post-training backdoor removal in FL. During training, clients identify suspicious samples using spectral analysis and activation clustering, then train lightweight Wasserstein Generative Adversarial Network with Gradient Penalty (WGAN-GP) models to capture trigger-related distributions. The generator parameters are aggregated server-side to construct a global representation of suspicious patterns without exposing raw data. After convergence, the server synthesizes trigger-approximating samples and applies machine unlearning to erase the trigger-target association by redistributing predictions toward a uniform distribution. Experimental evaluations on CIFAR-10 and GTSRB across three attack types and up to 40\% malicious participation demonstrate that SCRUB-FL reduces the backdoor attack success rate to as low as 3.88\% while maintaining over 91\% normal task accuracy, outperforming state-of-the-art defenses without requiring prior trigger knowledge or a large clean proxy dataset at the server.
\end{abstract}

\begin{IEEEImpStatement}
Federated Learning (FL) is increasingly deployed in privacy-sensitive applications, yet backdoor attacks remain a major threat because malicious behaviors can persist in the global model after training. Existing post-training defenses often require large labeled reference datasets, prior trigger knowledge, or extensive model modification, limiting their effectiveness. This work demonstrates that effective post-training backdoor removal can be achieved without access to client data, prior trigger knowledge, or a large clean server-side dataset. More importantly, it shows that information collected during distributed training can be leveraged to support the mitigation of backdoor attacks after model convergence. Experimental results indicate that backdoor attack success rates can be reduced below 3.88\% while maintaining over 91\% clean-task accuracy. The proposed solution contributes to the emerging intersection of federated learning, artificial intelligence (AI) security, and machine unlearning by providing a practical mechanism for removing adversarially embedded behaviors from trained models.
\end{IEEEImpStatement}


\section{Introduction}
\label{intro}

Federated Learning (FL) is a decentralized paradigm that enables multiple clients to collaboratively train a shared global model without exchanging raw data \cite{mcmahan2017communication}. Clients perform local training on private datasets and transmit only model updates to a central server for aggregation, preserving data locality. Despite this advantage, the decentralized and partially trusted structure of FL introduces serious security vulnerabilities, since the server has no direct access to the client-side datasets, the system is inherently susceptible to adversarial exploitation. Among these threats, backdoor attacks have emerged as one of the most critical and difficult-to-address \cite{feng2025survey}.

In a backdoor attack, malicious clients poison their local training data by embedding carefully crafted hidden triggers, such as pixel-level patches, frequency-domain perturbations, or semantic feature modifications, into a subset of samples while relabeling them to an attacker-designated target class. The resulting global model operates normally under standard inputs, successfully passing routine accuracy evaluations, yet produces attacker-controlled predictions whenever the trigger pattern is present at inference time. The severity of this threat is formally recognized by MITRE ATLAS (Adversarial  Threat Landscape for Artificial-Intelligence Systems), which catalogues backdoor attacks on machine learning models as a high-priority adversarial tactic under technique AML.T0018, explicitly highlighting their applicability to collaborative and federated learning pipelines~\cite{mitreatlas}. This institutional recognition underscores that backdoor attacks are not merely a theoretical concern but an actively tracked, real-world threat to deployed AI systems. This stealthy coexistence of normal functionality and embedded malicious behavior makes backdoor attacks particularly dangerous and resistant to detection through conventional model monitoring methods.

Existing defense strategies against backdoor attacks in FL can be broadly categorized into three classes: pre-training, in-training, and post-training defenses. Pre-training defenses sanitize or inspect client data before federated training begins. However, they are inherently static, relying on prior knowledge of known attack patterns and applying filtering only once before training commences, making them ineffective against adaptive or evolving backdoor strategies that emerge dynamically across training rounds. In-training defenses, such as robust aggregation~\cite{cao2020fltrust} and anomaly-based filtering~\cite{nguyen2022flame}, identify and downweight suspicious updates during aggregation. While promising, they remain vulnerable to sophisticated types of attacks, tend to destabilize model convergence, degrade normal task accuracy due to the exclusion of legitimate updates, and cannot guarantee the complete elimination of backdoor patterns that gradually accumulate across training rounds. Post-training defenses operate on the converged global model but typically rely on a limited clean proxy dataset at the server and lack direct knowledge of trigger patterns or poisoned clients after convergence. Consequently, they often fail to fully eliminate backdoor behaviors, either leaving residual malicious patterns through incomplete sanitization or causing unnecessary accuracy degradation through aggressive neuron pruning, a problem known as \emph{neuron entanglement}, wherein backdoor-sensitive neurons also encode legitimate semantic features, making their removal harmful to normal task performance. Therefore, despite the progress achieved by existing defenses, effective post-training sanitization in federated learning remains challenging due to the lack of sufficient knowledge about trigger patterns and malicious behaviors needed to act effectively in a federated setting. These challenges are further compounded by the trade-off between backdoor removal and accuracy preservation, the diversity of attack types considered in the defense literature, and the need for extensive retraining or fine-tuning after convergence.

To bridge this critical gap, we propose \textbf{SCRUB-FL} (\textbf{S}anitizing and \textbf{C}leansing \textbf{R}epresentations via \textbf{U}nlearning of \textbf{B}ackdoors), a novel and privacy-compliant framework for detecting and neutralizing backdoor attacks in a converged federated model. Rather than intervening before or during aggregation, SCRUB-FL operates after training has been accomplished, targeting the residual malicious behaviors embedded in the global model's parameters. The key insight driving our approach is that backdoor triggers induce characteristic activation anomalies in specific neurons, which can be exposed by feeding the model with synthetically generated suspicious samples that approximate the trigger distribution. To produce these samples without accessing raw client data, each client trains a lightweight Generative Adversarial  Network (GAN) on a locally extracted subset of suspicious samples, identified through spectral signature analysis and per class activation clustering. These complementary techniques enable the detection of both dominant poisoning directions in the feature space and anomalous neuron activation patterns that may not appear as strong spectral outliers. SCRUB-FL assumes a protocol-compliant adversary commonly adopted in federated learning security literature. Although malicious clients may poison local training data, they continue to follow the prescribed training protocol, including the lightweight GAN training step, to avoid detection and exclusion from the federation. The resulting generator parameters are communicated alongside model updates and aggregated server-side to form a global generative model. This global generator is then used to synthesize suspicious inputs, which are passed through the global classifier. Rather than pruning neurons that may encode both backdoor and legitimate features, SCRUB-FL applies machine unlearning, where the model is trained to produce a uniform, non-committal output distribution over all classes when presented with trigger-approximating inputs, while simultaneously preserving correct predictions on clean reference samples. This simultaneous forgetting preservation objective, with an amplified unlearning signal, erases the backdoor mapping without any permanent structural modification to the network, producing a sanitized model that retains high utility on benign tasks.

The main contributions of this work are summarized as follows: 

\begin{enumerate}

    \item Proposing SCRUB-FL, a post-training backdoor sanitization framework for federated learning that operates on the converged global model without requiring access to raw client data or a large labelled clean proxy dataset at the server.
    \item Designing a privacy-preserving suspicious pattern extraction pipeline in which each client identifies locally anomalous samples using spectral signatures and per-class clustering, and trains a      lightweight GAN to capture their distributional characteristics.
    \item Developing a server-side machine unlearning mechanism that synthesizes trigger-approximating samples from the aggregated global generator and applies a simultaneous forgetting-preservation objective with an amplified unlearning signal to erase trigger-to-target mappings, avoiding the neuron entanglement problem that causes accuracy degradation in pruning-based post-training defenses.
    \item Demonstrating through extensive experiments on CIFAR-10 and GTSRB that SCRUB-FL effectively suppresses backdoor attack success rates to as low as 3.88\% across diverse attack scenarios, including One-to-One, One-to-N, and N-to-One attacks under up to 40\% malicious participation, while maintaining high normal task accuracy, outperforming both aggregation-phase and post-training defens baselines.

\end{enumerate}

The remainder of this paper is structured as follows. Section~\ref{sec:related}  reviews related work on backdoor attacks and defense mechanisms in  federated learning. Section~\ref{sec:prelim} introduces the  background concepts underpinning our framework.  Section~\ref{sec:problem} formally defines the backdoor threat model  considered in this work. Section~\ref{sec:method} presents the  SCRUB-FL architecture and its core components.  Section~\ref{sec:experiments} reports experimental results and  comparative analysis. Finally, Section~\ref{sec:conclusion} concludes  the paper and outlines directions for future research. 

\section{Related Work}
\label{sec:related}

In this section, we review representative works across these phases, examine their methodological contributions and limitations, and identify the gaps that motivate the design of SCRUB-FL.

The authors of~\cite{wang2025resisting} propose a defense where each client searches for a path in the model weight space between the received global model and its own locally trained model, selecting an intermediate point that reduces backdoor influence while preserving normal task performance. However, this process is computationally expensive for edge devices and forces a trade-off between security and accuracy, with the defense breaking down when most clients are malicious. To move detection to the server side, the authors of~\cite{alharbi2025robust} propose inspecting client gradients directly using a local reference dataset, with the detection logic hidden to prevent attackers from bypassing it. While promising, the method produces a high number of false alarms, up to 51.5\%, when the reference dataset is small, and assumes that attackers cannot reverse-engineer the hidden detection code, which may not hold against advanced adversaries. The authors of~\cite{tan2025defending} propose a server-side defense that uses a reinforcement learning agent to assign trust scores to client updates by comparing them against a reference built from a small clean dataset. Although the method adapts well to moderate attack scenarios, it requires the server to hold clean data, which conflicts with the privacy goals of federated learning, and its performance drops significantly when the number of malicious clients is high. The authors of~\cite{obioma2025defending} use clustering to group client updates and select the most trustworthy ones, then apply knowledge distillation to improve the global model using an unlabeled dataset on the server. Despite its effectiveness, the method requires a large amount of server-side data and runs clustering and distillation every training round, making it too costly for large-scale federated systems. The authors of~\cite{bi2025securing} detect suspicious client updates by checking whether their outputs fall outside the normal data distribution, then apply privacy noise to further reduce backdoor influence. However, the added noise consistently hurts normal task accuracy, and the method fails against attacks that are designed to look like normal data and thus avoid detection. The authors of~\cite{zhang2024flpurifier} evaluate each client's behavior across multiple metrics such as how much their updates align with the global model direction and how consistent they are across rounds, filtering out updates that appear coordinated or abnormally large. Despite covering many attack patterns, the method needs a clean reference dataset on the server and works poorly on small or simple network architectures. The authors of~\cite{wu2024unlearning} split the model into two parts: a feature extractor trained using self-supervised learning to avoid linking trigger patterns to target labels, and a classifier aggregated by the server. While effective, this design requires all clients to change how they train their models, which adds significant cost and slows down learning in the early training rounds. Several works have also focused on cleaning the global model after training is complete. The authors of~\cite{huang2025fedcleanse} ask clients to vote on which neurons are most important for the main task, and the server uses these votes to suppress neurons that may be linked to backdoor behavior. However, suppressing too many neurons accidentally removes useful ones, hurting accuracy and making the model vulnerable to attacks that spread the backdoor across many neurons to avoid detection. The authors of~\cite{wang2026shift} remove a known attacker's past updates from the global model by subtracting them directly, then use knowledge distillation to recover normal performance. The main drawback is that the server must already know which clients were malicious and must have access to a clean dataset for distillation, both of which are strong assumptions that are hard to satisfy in practice. The authors of~\cite{zhu2023adfl} limit the size of neuron activations in the trained global model to cut off the unusually large responses caused by poisoned inputs. While this confirms that acting on the model after training can work, the method depends on high-quality synthetic data to identify the right thresholds, and poor-quality data leads to either over-pruning or leaving the backdoor in place. The authors of~\cite{walter2024mitigating} train a GAN on the server to generate samples that reflect backdoor patterns in the global model, relabel them using a clean reference model, and use knowledge distillation to remove the backdoor. However, the method requires the server to hold labeled clean data, which goes against the privacy principles of federated learning, and the quality of the defense depends entirely on whether the GAN successfully learns the backdoor pattern.

The works reviewed above reveal two clear gaps in the existing literature. First, most defenses focus on filtering or adjusting client updates during training, but they cannot remove backdoor patterns that have already been built into the model over many training rounds, and they often hurt normal accuracy or add too much computational cost. Second, defenses that act after training is complete typically rely on having clean data at the server, knowing which clients were malicious, or having some prior knowledge of the backdoor trigger, none of which can be assumed in a real federated learning setting. These limitations highlight the need for a post-training defense that works without any clean data at the server, without knowing what the trigger looks like, and without changing how clients train their models. This is exactly what SCRUB-FL aims to achieve by collecting information about suspicious patterns during training in a privacy-preserving way, and using it to erase the backdoor mapping through machine unlearning after training is done, without requiring neuron removal or structural modification to the network. 

\section{System Preliminaries}
\label{sec:prelim}

In this section, we introduce the core concepts and techniques that form the foundation of our proposed framework. Starting by describing the standard federated learning process, followed by an overview of Generative Adversarial Networks and their role in our approach, and conclude with a description of the two client-side anomaly detection techniques used to identify suspicious local samples.

For clarity and consistency throughout this paper, Table~\ref{tab:symbols}
summarizes the key symbols and notation used in our formulation.
 
\begin{table}[t]
\caption{Summary of Key Symbols and Notation}
\label{tab:symbols}
\centering
\renewcommand{\arraystretch}{1.0}
\begin{tabular}{p{0.18\columnwidth} p{0.75\columnwidth}}
\toprule
\textbf{Symbol} & \textbf{Description} \\
\midrule
$N$ & Total number of clients in the federation \\
$T$ & Total number of federated training rounds \\
$t$ & Index of the current training round \\
$\mathcal{D}_i$ & Local dataset of client $i$ \\
$\mathcal{D}_i^{s}$ & Suspicious sample subset extracted by client $i$ \\
$\mathcal{D}_i^{p}$ & Poisoned local dataset of a malicious client $i$ \\
$n_i$ & Number of samples in $\mathcal{D}_i$ \\
$\mathbf{x}$ & Input data sample \\
$y$ & True class label of $\mathbf{x}$ \\
$y^{*}$ & Attacker’s target class label \\
$\tau$ & Backdoor trigger pattern \\
$\mathbf{x}_\tau$ & Triggered sample, $\mathbf{x} \oplus \tau$ \\
$\rho$ & Poisoning ratio in $\mathcal{D}_i^{p}$ \\
$\mathcal{I}^{m}$ & Set of malicious client indices \\
$\mathcal{W}$ & Global model parameters \\
$\mathcal{W}_i^{t}$ & Local model of client $i$ at round $t$ \\
$\mathcal{W}^{t}$ & Global model at round $t$ \\
$\eta$ & Local learning rate \\
$\mathcal{L}$ & Training loss function \\
$G_{\phi}$ & Generator with parameters $\phi$ \\
$D_{\psi}$ & Discriminator with parameters $\psi$ \\
$G_{\phi}^{t}$ & Local generator of client $i$ at round $t$ \\
$\bar{G}_{\phi}$ & Aggregated global generator \\
$\mathbf{x}_g$ & Generated suspicious sample \\
$f_{\mathcal{W}}(\cdot)$ & Global classifier \\
$u$ & Uniform soft label vector, $\frac{1}{C}\mathbf{1} \in \mathbb{R}^{C}$ \\
$\lambda_{u}$ & Unlearning amplification multiplier \\
$\eta_{u}$ & Unlearning learning rate \\
$E_{u}$ & Number of unlearning epochs \\
$\mathcal{L}_{\text{forget}}$ & Forgetting loss over generated suspicious samples \\
$\mathcal{L}_{\text{preserve}}$ & Preservation loss over clean reference inputs \\
$\mathcal{L}_{\text{unlearn}}$ & Combined unlearning objective \\
$\hat{W}$ & Sanitized global model parameters \\
\bottomrule
\end{tabular}
\end{table}

\subsection{Federated Learning}

Federated learning is a distributed machine learning framework that allows a group of $N$ clients to jointly train a shared global model without sharing their raw data~\cite{mcmahan2017communication}. Each client $i$ holds a private local dataset $\mathcal{D}_i = \{(\mathbf{x}_j, y_j)\}_{j=1}^{n_i}$, where $\mathbf{x}_j$ denotes an input sample and $y_j$ its corresponding class label.

Training proceeds over $T$ communication rounds. At the start of each round $t$, the server broadcasts the current global model parameters $\mathcal{W}^{t}$ to all participating clients. Each client $i$ then performs local training by minimizing the following local objective:

\begin{equation}
    \mathcal{W}_i^{t} = \mathcal{W}^{t} - \eta \, \nabla \mathcal{L}\!\left(\mathcal{W}^{t};\, \mathcal{D}_i\right) 
\label{eq:local_update}
\end{equation}

where $\eta$ is the local learning rate and $\mathcal{L}$ is the training loss function. Once local training is complete, each client sends its updated parameters $\mathcal{W}_i^{t}$ back to the server. The server aggregates all received updates using a weighted average to produce the next global model:

\begin{equation}
    \mathcal{W}^{t+1} = \sum_{i=1}^{N} \frac{n_i}{\sum_{j=1}^{N} n_j} \, \mathcal{W}_i^{t}
\label{eq:aggregation} 
\end{equation}

This process repeats for $T$ rounds until the global model converges. The key privacy property of this framework is that raw data $\mathcal{D}_i$ never leaves the client device; only model parameters are communicated.

\subsection{Generative Adversarial Networks}

A Generative Adversarial Network (GAN) is a generative model composed of two neural networks that are trained together through a competitive process~\cite{goodfellow2014generative}. The first network, called the generator $G_{\phi}$, takes a random noise vector $\mathbf{z} \sim p(\mathbf{z})$ as input and produces a synthetic sample $G_{\phi}(\mathbf{z})$ that resembles samples from a target data distribution. The second network, called the discriminator $D_{\psi}$, receives either a real sample from the target distribution or a synthetic sample from the generator, and tries to distinguish between the two.

The two networks are trained simultaneously by optimizing the following minimax objective:

\begin{equation}
    \min_{\phi} \max_{\psi} \; \mathbb{E}_{\mathbf{x} \sim p_{\text{data}}} \left[\log D_{\psi}(\mathbf{x})\right] + \mathbb{E}_{\mathbf{z} \sim p(\mathbf{z})} \left[\log \left(1 - D_{\psi}(G_{\phi}(\mathbf{z}))\right)\right]
\label{eq:gan_objective}
\end{equation}

As training progresses, the generator learns to produce increasingly realistic samples that the discriminator can no longer distinguish from real ones. In SCRUB-FL, we use a Wasserstein GAN with Gradient Penalty (WGAN-GP)~\cite{gulrajani2017improved}, a more stable variant of the standard GAN that replaces the original loss with a Wasserstein distance-based objective:

\begin{equation}
    \min_{\phi} \max_{\|\psi\|_L \leq 1} \; \mathbb{E}_{\mathbf{x} \sim p_{\text{data}}} \left[D_{\psi}(\mathbf{x})\right] - \mathbb{E}_{\mathbf{z} \sim p(\mathbf{z})} \left[D_{\psi}(G_{\phi}(\mathbf{z}))\right]
\label{eq:wgan_objective}
\end{equation}

where the discriminator $D_{\psi}$ is constrained to be a 1-Lipschitz function enforced via gradient penalty. This formulation provides more stable training gradients and converges reliably even when trained on small datasets, making it well-suited for learning from the limited suspicious subsets available at each client.

\subsection{Spectral Signature Analysis}

Spectral signature analysis is a technique used to identify potentially poisoned samples within a dataset by examining the statistical structure of their feature representations~\cite{chen2018detecting}. The core idea is that backdoor triggers introduce a shared consistent feature across poisoned samples. As the model learns to associate this trigger with the attacker’s target label, the latent representations of poisoned samples become more aligned with one another than with clean samples from the same class. This creates a dominant outlier direction in the feature covariance structure, making poisoned samples statistically distinguishable through spectral analysis.

Given the feature representations of all samples in a class, extracted from an intermediate layer of the model, we compute the covariance matrix of these representations and apply Singular Value Decomposition (SVD) to identify the top principal directions of variation. For each sample $\mathbf{x}_j$, its spectral signature score is computed as its projection onto the top singular vector $\mathbf{v}_1$:

\begin{equation}
    s_j = \left( \mathbf{h}_j - \bar{\mathbf{h}} \right)^{\top} \mathbf{v}_1
\label{eq:spectral_score}
\end{equation}

where $\mathbf{h}_j$ is the feature representation of sample $\mathbf{x}_j$ and $\bar{\mathbf{h}}$ is the mean feature representation of all samples in the class. Samples with spectral scores that fall significantly above the mean are flagged as potentially poisoned, as they align strongly with the dominant outlier direction introduced by the trigger. In SCRUB-FL, this analysis is performed locally on each client’s dataset to extract the suspicious subset $\mathcal{D}_i^{s}$ before GAN training begins.

\subsection{Activation Clustering}

Activation clustering is a complementary anomaly detection technique that identifies suspicious samples by analyzing how they group in the activation space of a neural network~\cite{chen2018detecting}. The intuition behind adopting this technique into our solution is that poisoned samples, despite carrying the same label as clean samples of the target class, activate a distinct set of neurons due to the presence of the trigger, causing them to form a separate cluster in the model's internal representation space.

Given a set of samples from class $c$, we extract their activation vectors $\{\mathbf{a}(\mathbf{x}_j)\}$ from a target layer of the local model and apply a clustering algorithm such as K-Means to partition them into two groups. The smaller or more isolated cluster is considered the suspicious group, as it corresponds to samples whose internal representations deviate from the majority of clean samples in that class. Formally, for each class $c$, we partition the activation vectors into two clusters $\mathcal{C}_1^c$ and $\mathcal{C}_2^c$ by solving:

\begin{equation}
    \underset{\mathcal{C}_1^c,\, \mathcal{C}_2^c}{\arg\min} \sum_{k \in \{1,2\}} \sum_{\mathbf{a}_j \in \mathcal{C}_k^c} \left\| \mathbf{a}_j - \boldsymbol{\mu}_k^c \right\|^2
\label{eq:kmeans}
\end{equation}

where $\boldsymbol{\mu}_k^c$ is the centroid of cluster $k$ for class $c$. The cluster with fewer members is taken as the suspicious subset for that class. In SCRUB-FL, activation clustering is applied in parallel with spectral signature analysis, and the union of samples flagged by both methods forms the final suspicious subset $\mathcal{D}_i^{s}$ used to train the local GAN. 

\section{Problem Formulation}
\label{sec:problem}

In this section, we formally define the federated learning setting considered in this work, describe the backdoor attack models that SCRUB-FL is designed to defend against, and specify the capabilities and limitations assumed for both the attacker and the defender. A clear threat model is essential to understanding the scope of our defense and the conditions under which it is expected to operate.

\subsection{Federated Learning Setting}

We consider a standard federated learning system consisting of one central server and $N$ clients. Each client $i$ holds a private local dataset $\mathcal{D}_i$ that is never shared with the server or with other clients. Training proceeds over $T$ communication rounds as described in Section~\ref{sec:prelim}. Among the $N$ participating clients, a subset $\mathcal{I}^{m} \subset \{1, \ldots, N\}$ are malicious, meaning they actively attempt to embed backdoor behaviors into the global model during local training. The remaining clients $\{1, \ldots, N\} \setminus \mathcal{I}^{m}$ are honest and train on clean, unmodified local datasets. We assume that the fraction of malicious clients satisfies $|\mathcal{I}^{m}| / N < 0.5$, meaning that honest clients constitute the majority of participants, which is a standard and widely adopted assumption in the federated learning security literature~\cite{wang2024mudguard}. 

\subsection{Backdoor Attack Model}

A backdoor attack aims to embed a hidden malicious behavior into the global model such that the model performs normally on clean inputs but produces attacker-controlled predictions whenever a specific trigger pattern is present in the input. We first define the poisoning process used by malicious clients, followed by the different backdoor attack types considered in this work.

\subsubsection{Data Poisoning}
Each malicious client $i \in \mathcal{I}^{m}$ constructs a poisoned local dataset $\mathcal{D}_i^{p}$ by injecting trigger-bearing samples into a fraction $\rho$ of its clean data. Formally, the poisoned dataset is defined as:

\begin{equation}
    \mathcal{D}_i^{p} = \mathcal{D}_i^{\text{clean}} \cup \mathcal{D}_i^{\text{trigger}}
\label{eq:poisoned_dataset}
\end{equation}

where $\mathcal{D}_i^{\text{clean}}$ contains the unmodified samples and $\mathcal{D}_i^{\text{trigger}}$ contains the poisoned samples, with $|\mathcal{D}_i^{\text{trigger}}| = \rho \cdot |\mathcal{D}_i|$. Each poisoned sample is constructed by applying a trigger pattern $\tau$ to a clean input $\mathbf{x}$ and replacing its true label $y$ with the attacker's target label $y^{*}$:

\begin{equation}
    \mathcal{D}_i^{\text{trigger}} = \left\{ \left(\mathbf{x} \oplus \tau,\; y^{*}\right) : \left(\mathbf{x}, y\right) \in \mathcal{D}_i,\; y \neq y^{*} \right\}
\label{eq:trigger_injection}
\end{equation}

where $\oplus$ denotes the trigger application operation, which may represent pixel-level overlay, frequency-domain perturbation, or any other method of embedding the trigger into the input sample.

\subsubsection{Attack Types}

Depending on how the trigger is structured and how many target behaviors are involved, we distinguish between the following three attack types, all of which SCRUB-FL is designed to handle:

\textbf{One-to-One Attack.} A single trigger $\tau$ is applied to samples from one or more source classes to cause the model to predict a single fixed target label $y^{*}$. This is the most common and well-studied form of backdoor attack:

\begin{equation}
    f_{\mathcal{W}}(\mathbf{x} \oplus \tau) = y^{*}, \quad \forall\; (\mathbf{x}, y) \in \mathcal{D}_i,\; y \neq y^{*}
\label{eq:one_to_one}
\end{equation}

\textbf{One-to-N Attack.} Multiple trigger variants $\{\tau_1, \tau_2, \ldots, \tau_K\}$ are used, each mapping triggered inputs to a different target label. This makes detection harder as the attack spans multiple output behaviors:

\begin{equation} f_{\mathcal{W}}(\mathbf{x} \oplus \tau_k) = y_k^{*}, \quad k = 1, 2, \ldots, K
\label{eq:one_to_n}
\end{equation}

\textbf{N-to-One Attack.} Multiple triggers $\{\tau_1, \tau_2, \ldots, \tau_K\}$ must all be present simultaneously in the input to activate the backdoor. No individual trigger alone causes a misclassification, making this attack particularly difficult to detect:

\begin{equation}
    f_{\mathcal{W}}\!\left(\mathbf{x} \oplus \bigoplus_{k=1}^{K} \tau_k \right) = y^{*}
\label{eq:n_to_one}
\end{equation}

\subsubsection{Attack Objective}

The objective of the attacker is to craft local model updates that, when aggregated with honest client updates, cause the global model to satisfy two conditions simultaneously. First, the model must behave normally on clean inputs, maintaining high classification accuracy on the main task to avoid basic detection. Second, the model must produce the target label $y^{*}$ whenever the trigger is present. This dual objective can be expressed as the following local training problem solved by each malicious client:

\begin{equation}
    \min_{\mathcal{W}_i^{t}} \; \mathcal{L}\!\left(\mathcal{W}_i^{t};\, \mathcal{D}_i^{\text{clean}}\right) + \lambda_a \cdot \mathcal{L}\!\left(\mathcal{W}_i^{t};\, \mathcal{D}_i^{\text{trigger}}\right)
\label{eq:attacker_objective}
\end{equation}

where $\lambda_a > 0$ is a scaling factor that controls the relative weight given to the backdoor task versus the main task during malicious local training. A higher $\lambda_a$ produces stronger backdoor injection but may increase the detectability of the malicious update.

\subsection{Attacker Capabilities and Assumptions}

The attacker is assumed to have control over a subset of compromised clients $\mathcal{I}^{m}$, including the ability to modify local training data (e.g., via data poisoning). The attacker can also coordinate across malicious clients to distribute the backdoor trigger and improve its effectiveness while reducing per-client detectability. However, the attacker has no access to the server-side aggregation process, cannot directly modify the global model, and has no prior knowledge of the server’s defense mechanism. Furthermore, the attacker cannot observe or interfere with the local training of honest clients. These assumptions reflect a realistic and widely adopted threat model in federated learning backdoor attack literature, where malicious clients are compromised while the server and benign clients remain trusted~\cite{feng2025survey, nguyen2022flame}. The client has no visibility into which samples are flagged, and therefore has no mechanism to selectively manipulate the composition of the subset used to train the local WGAN-GP. The attacker's control ends at data submission; everything downstream is handled internally by the pipeline. The adversary is further assumed to be protocol-compliant during training. Malicious clients can manipulate their local data but cannot alter the server-mandated training process or directly modify the global aggregation procedure. To avoid detection and exclusion, compromised clients follow the training protocol, including the GAN training step, which appears as a standard task. 

\subsection{Defender Capabilities and Assumptions}

The server performs standard federated aggregation as defined in Equation~(\ref{eq:aggregation}) and has access to all client model updates $\{\mathcal{W}_i^{t}\}_{i=1}^{N}$ in each round. Critically, the server has no access to any client's raw data $\mathcal{D}_i$, no prior knowledge of the trigger pattern $\tau$, a small labeled clean proxy dataset, and no information about which clients belong to $\mathcal{I}^{m}$. These constraints reflect the strict privacy and trust assumptions that define the federated learning setting and distinguish our problem from centralized backdoor defense scenarios. The server does, however, receive the generator parameters $G_{\phi_i}^{t}$ from each client alongside the classifier updates, as described in Section~\ref{sec:method}, which are the only additional information required by SCRUB-FL beyond the standard federated protocol.

\subsection{Defense Goal}

Given the federated setting and threat model described above, the goal of SCRUB-FL is to produce a sanitized global model $\hat{\mathcal{W}}$ after training has converged such that the following two conditions hold:

\textbf{Backdoor Elimination.} The sanitized model no longer produces the attacker's target label when presented with triggered inputs. Formally, for all attack types defined in Equations~(\ref{eq:one_to_one}--\ref{eq:n_to_one}):

\begin{equation}
    f_{\hat{\mathcal{W}}}(\mathbf{x} \oplus \tau) \neq y^{*}, \quad \forall\; (\mathbf{x}, y) \in \mathcal{D}^{\text{test}}
\label{eq:backdoor_elimination}
\end{equation}

\textbf{Utility Preservation.} The sanitized model maintains high classification accuracy on clean inputs, comparable to the performance of a model trained without any malicious participants:

\begin{equation}
    \left| \text{Acc}\!\left(f_{\hat{\mathcal{W}}}\right) - \text{Acc}\!\left(f_{\mathcal{W}^*}\right) \right| \leq \epsilon
\label{eq:utility_preservation}
\end{equation}

where $f_{\mathcal{W}^*}$ denotes an ideally trained clean global model and $\epsilon > 0$ is a small acceptable accuracy tolerance. Achieving both conditions simultaneously, without access to clean server-side data or knowledge of the trigger, is the central challenge that SCRUB-FL is designed to address. 


\begin{figure*}[!t]
    \centering
    \includegraphics[width=0.90\textwidth]{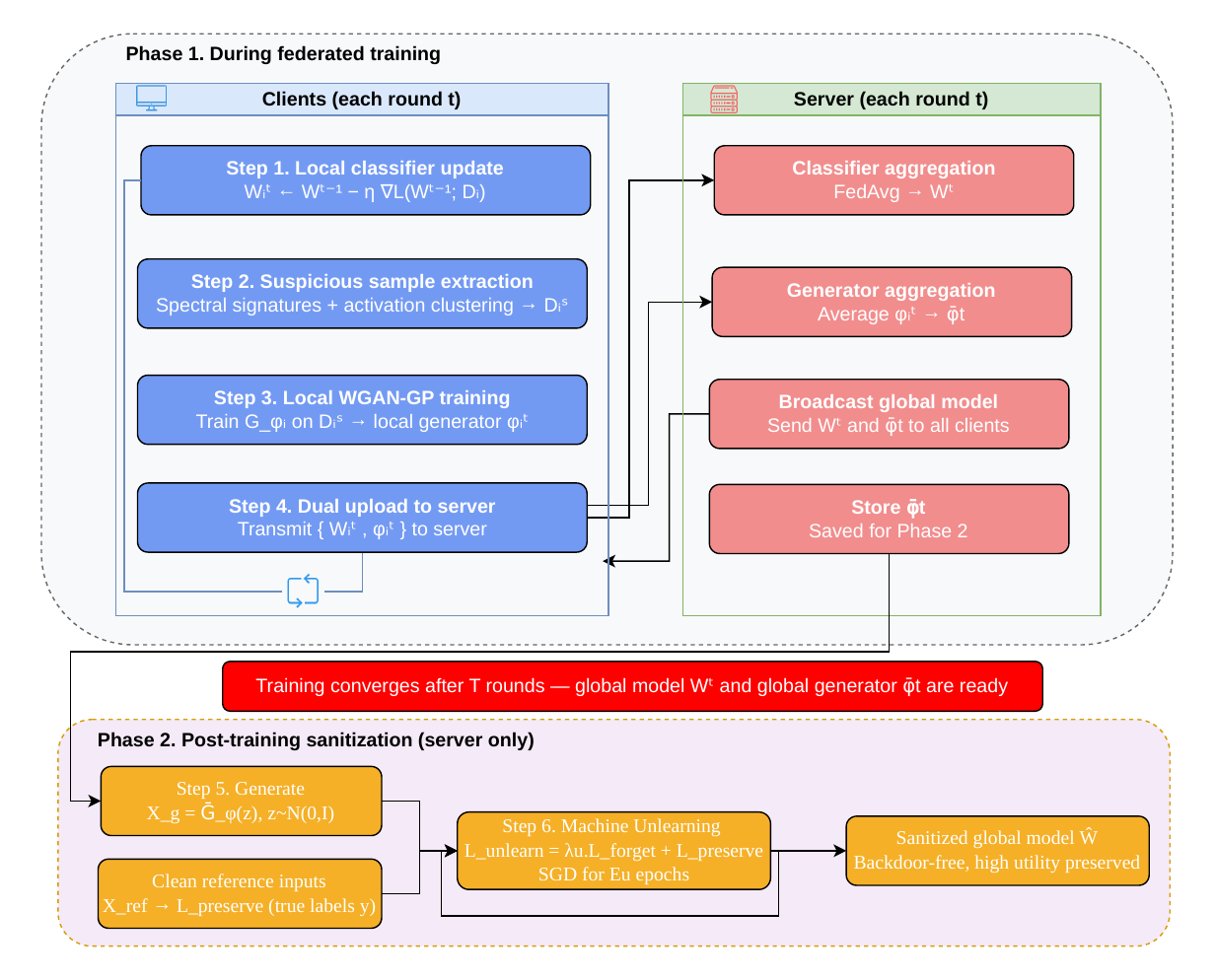}
    \caption{Overview of the SCRUB-FL framework, illustrating the federated training phase with local anomaly detection and GAN-based modeling, followed by post-training server-side sanitization via machine unlearning.}
    \label{fig:architecture}
\end{figure*}

\section{The SCRUB-FL Framework}
\label{sec:method}

In this section, we present the full architecture of SCRUB-FL (Sanitizing and Cleansing Representations via Unlearning of Backdoors). SCRUB-FL operates in two sequential phases that together address the knowledge gap described in Section~\ref{sec:problem}. The first phase runs concurrently with the standard federated training process and is responsible for capturing a compact, privacy-preserving representation of suspicious pattern distributions at each client. The second phase executes once after the global model has fully converged. It uses the captured knowledge to identify and neutralize backdoor-sensitive neurons in the global model.

Figure~\ref{fig:architecture} provides a high-level overview of the SCRUB-FL architecture, illustrating how the two phases interact with the standard federated pipeline. The remainder of this section describes each component of the framework in detail, following the order in which they execute during a full training run.

\subsection{Phase 1. Suspicious Pattern Capture During Training}

The first phase of SCRUB-FL augments the standard federated training pipeline with three additional steps performed locally at each client before transmitting updates to the server: suspicious sample extraction, local GAN training, and dual update transmission. These steps add no burden to the server and require no changes to the aggregation protocol beyond receiving the additional generator parameters.

\subsubsection{Step 1. Local Classifier Update} At each training round $t$, each client $i$ performs standard local training by minimizing Equation~(\ref{eq:local_update}), producing updated parameters $\mathcal{W}_i^t$ that are used both for the subsequent detection step, and for transmission to the server.

\subsubsection{Step 2. Local Suspicious Sample Extraction}  At each training round $t$, after performing the standard local model update defined in Equation~(\ref{eq:local_update}), a locally executed anomaly detection routine is applied on
each client $i$ runs to identify a subset of its data $\mathcal{D}_i^{s} \subseteq \mathcal{D}_i$ that may contain trigger-embedded samples. This is done using two complementary techniques: spectral signature analysis and activation clustering, both introduced in Section~\ref{sec:prelim}. Since this extraction runs as an internal pipeline step on the client's submitted dataset, the client, whether honest or malicious, has no mechanism to observe or influence which samples are flagged.

Spectral signature analysis is applied per class. For each class $c$ in client $i$'s local dataset, the client extracts the feature representations $\{\mathbf{h}_j\}$ of all samples belonging to class $c$ from an intermediate layer of its local model $f_{\mathcal{W}_i^{t}}$. It then computes the spectral score of each sample using Equation~(\ref{eq:spectral_score}) and flags samples whose scores exceed a class-specific threshold as suspicious:

\begin{equation}
    \mathcal{S}_i^{\text{spec},c} = \left\{ \mathbf{x}_j \in \mathcal{D}_i^c \;:\; s_j > \bar{s}^c + \kappa \cdot \sigma_{s}^c \right\}
\label{eq:spectral_threshold}
\end{equation}

where $\bar{s}^c$ and $\sigma_{s}^c$ are the mean and standard deviation of the spectral scores within class $c$, and $\kappa > 0$ is a sensitivity parameter that controls how aggressively samples are flagged.

In parallel, activation clustering is applied by extracting the activation vectors $\{\mathbf{a}(\mathbf{x}_j)\}$ of all samples in class $c$ from the same intermediate layer and partitioning them into two clusters using K-Means as defined in Equation~(\ref{eq:kmeans}). The smaller cluster is taken as the suspicious group for that class:
\begin{equation}
    \mathcal{S}_i^{\text{clus},c} = \arg\min_{k \in \{1,2\}} \left| \mathcal{C}_k^c \right|
\label{eq:suspicious_cluster}
\end{equation}

The final suspicious subset for client $i$ is formed by taking the union of flagged samples across both methods and all classes:

\begin{equation}
    \mathcal{D}_i^{s} = \bigcup_{c} \left( \mathcal{S}_i^{\text{spec},c} \cup \mathcal{S}_i^{\text{clus},c} \right)
\label{eq:suspicious_union}
\end{equation}

Using the union of both methods ensures that trigger-bearing samples missed by one technique are still captured by the other, improving the coverage of the suspicious subset without requiring either method to work perfectly on its own.

\subsubsection{Step 3. Local GAN Training on Suspicious Samples}  Once the suspicious subset $\mathcal{D}_i^{s}$ has been identified, client $i$ trains a lightweight WGAN-GP on this subset to learn the underlying distribution of potential trigger patterns. The generator $G_{\phi_i}$ takes a random noise vector $\mathbf{z} \sim \mathcal{N}(\mathbf{0}, \mathbf{I})$ as input and produces a synthetic sample $G_{\phi_i}(\mathbf{z})$ that resembles samples in $\mathcal{D}_i^{s}$. The discriminator $D_{\psi_i}$ is trained to distinguish between real samples from $\mathcal{D}_i^{s}$ and synthetic samples from $G_{\phi_i}$.  Following the WGAN-GP formulation introduced in Equation~(\ref{eq:wgan_objective}), the local training objective for client $i$'s GAN is:

\begin{equation}
\begin{aligned}
\min_{\phi_i} \max_{\psi_i} \; 
& \mathbb{E}_{\mathbf{x} \sim \mathcal{D}_i^{s}} \left[ D_{\psi_i}(\mathbf{x}) \right] \\
& - \mathbb{E}_{\mathbf{z} \sim \mathcal{N}(\mathbf{0},\mathbf{I})} 
\left[ D_{\psi_i}(G_{\phi_i}(\mathbf{z})) \right] \\
& + \lambda_{\text{gp}} \cdot \mathbb{E}_{\hat{\mathbf{x}}} 
\left[ \left( \left\| \nabla_{\hat{\mathbf{x}}} D_{\psi_i}(\hat{\mathbf{x}}) \right\|_2 - 1 \right)^2 \right]
\end{aligned}
\label{eq:local_wgan}
\end{equation}

where $\hat{\mathbf{x}}$ is sampled uniformly along straight lines between real and generated samples, and $\lambda_{\text{gp}} > 0$ is the gradient penalty coefficient. Once training converges, the generator $G_{\phi_i}$ encodes the distributional characteristics of the suspicious samples in $\mathcal{D}_i^{s}$ without storing or transmitting any raw samples, thereby preserving data privacy.

To keep the computational overhead manageable, the generator architecture is intentionally kept lightweight, consisting of a small number of transposed convolutional layers followed by a batch normalization and activation layer. The GAN is trained for a fixed number of local epochs $E_g$, which is a hyperparameter set independently of the main classifier training epochs $E$.

\subsubsection{Step 4. Dual Update Transmission}  After completing both local classifier training and local GAN training, client $i$ transmits two sets of parameters to the server at the end of round $t$:

\begin{equation}
    \text{Upload}_i^{t} = \left\{ \mathcal{W}_i^{t},\; \phi_i^{t} \right\}
\label{eq:dual_upload}
\end{equation}

where $\mathcal{W}_i^{t}$ are the standard local classifier parameters and $\phi_i^{t}$ are the local generator parameters. The discriminator parameters $\psi_i^{t}$ are discarded after local training as they are not needed server-side. This keeps the additional communication overhead limited to the size of the generator, which is significantly smaller than the main classifier.

\subsection{Server-Side Dual Aggregation}

Upon receiving the uploads from all $N$ clients at the end of round $t$, the server performs two separate aggregation steps.

\subsubsection{Classifier Aggregation}  The classifier parameters are aggregated using the standard weighted average defined in Equation~(\ref{eq:aggregation}), producing the updated global classifier $\mathcal{W}^{t+1}$. This step is identical to standard FedAvg and requires no modification.

\subsubsection{Generator Aggregation}  The generator parameters from all clients are aggregated using a simple average to produce the global generator $\bar{G}_{\phi}$:
\begin{equation}
    \bar{\phi}^{t} = \frac{1}{N} \sum_{i=1}^{N} \phi_i^{t}
\label{eq:generator_aggregation}
\end{equation}

The resulting global generator $\bar{G}_{\phi}$ encodes the collective suspicious pattern knowledge from all participating clients. Since both honest and malicious clients contribute to this aggregation, the global generator captures a broad representation of patterns that appeared statistically anomalous across the federation. Crucially, the global generator is not used during training for any classification or aggregation decision. It is stored server-side and only activated after training has fully converged to support the post-training sanitization phase. 

\subsection{Phase 2. Post-Training Backdoor Sanitization
via Machine Unlearning}

Once the global classifier $f_{\mathcal{W}^T}$ has converged after $T$ rounds, SCRUB-FL enters its second phase. Rather than applying neuron pruning, which risks removing neurons that encode both backdoor and legitimate features, a phenomenon that is known as \emph{neuron entanglement}, SCRUB-FL employs machine unlearning to erase the trigger-to-target mapping through gradient-based optimization, preserving all network parameters while eliminating malicious behavior.

\subsubsection{Step 5. Trigger-Approximating Sample Generation}

The server uses the converged global generator
$\bar{G}_{\phi}^{T}$ to produce $M$ synthetic suspicious
samples:

\begin{equation}
    \mathcal{X}_g = \left\{
    x_g^{(m)} = \bar{G}_{\phi}^{T}(z^{(m)})
    : z^{(m)} \sim \mathcal{N}(\mathbf{0}, \mathbf{I}),
    m = 1, \ldots, M
    \right\}
    \label{eq:generated_samples}
\end{equation}

These samples approximate the distribution of
trigger-embedded inputs learned collectively from all clients
during training, and serve as probe inputs for the
unlearning phase.

\subsubsection{Step 6. Machine Unlearning via Trigger Distribution Flattening}

The unlearning objective operates on two simultaneous losses. The \emph{forgetting loss} trains the model to produce a uniform output distribution over all $C$ classes when presented with generated suspicious samples, erasing the specific trigger-to-target mapping without removing any neuron weights:

\begin{equation}
    \mathcal{L}_{\text{forget}}(\mathcal{W}) =
    \frac{1}{|\mathcal{X}_g|}
    \sum_{x_g \in \mathcal{X}_g}
    \mathcal{H}\!\left(
    f_{\mathcal{W}}(x_g),\; \mathbf{u}
    \right)
    \label{eq:forget_loss}
\end{equation}

where $\mathcal{H}$ denotes the cross-entropy function and
$\mathbf{u} = \frac{1}{C}\mathbf{1} \in \mathbb{R}^C$ is
a uniform soft label vector assigning equal probability
to all classes.

The \emph{preservation loss} ensures normal task accuracy
is maintained by training on clean reference samples with
their true labels:

\begin{equation}
    \mathcal{L}_{\text{preserve}}(\mathcal{W}) =
    \frac{1}{|\mathcal{X}_{\text{ref}}|}
    \sum_{(x,y) \in \mathcal{X}_{\text{ref}}}
    \mathcal{H}\!\left(f_{\mathcal{W}}(x),\; y\right)
    \label{eq:preserve_loss}
\end{equation}

To prevent the preservation loss from suppressing the
forgetting signal, the forgetting loss is amplified by
scalar multiplier $\lambda_u > 1$:

\begin{equation}
    \mathcal{L}_{\text{unlearn}}(\mathcal{W}) =
    \lambda_u \cdot \mathcal{L}_{\text{forget}}(\mathcal{W})
    + \mathcal{L}_{\text{preserve}}(\mathcal{W})
    \label{eq:unlearn_total}
\end{equation}

where $\lambda_u = 3.0$ is determined empirically to
balance forgetting speed against accuracy retention.
The model is updated via SGD for $E_u$ unlearning epochs:

\begin{equation}
    \hat{\mathcal{W}} \leftarrow \mathcal{W}^{T} -
    \eta_u \nabla_{\mathcal{W}}
    \mathcal{L}_{\text{unlearn}}(\mathcal{W}^{T})
    \label{eq:unlearn_update}
\end{equation}

where $\eta_u$ is the unlearning learning rate. This
ensures the model converges toward a state where triggered
inputs produce flat, non-committal outputs while clean
inputs retain their original classification behavior,
achieving both sanitization conditions stated in Equations~(\ref{eq:backdoor_elimination}) and~(\ref{eq:utility_preservation}) through the combined objective of Equations~(\ref{eq:forget_loss}--\ref{eq:unlearn_update}), without any permanent structural modification to the network.

\subsection{Complete SCRUB-FL Algorithm}

Algorithm~\ref{alg:flpbs} summarizes the complete execution
of SCRUB-FL across all $T$ federated training rounds and
the subsequent post-training sanitization phase.

The overall complexity of SCRUB-FL is dominated by three components, namely federated training, local GAN optimization, and post-training unlearning. During each communication round, the cost is $\mathcal{O}(N \cdot C_{\text{local}})$, where $C_{\text{local}}$ includes local model updates, spectral analysis, and activation clustering. The WGAN-GP training introduces an additional complexity of $\mathcal{O}(N \cdot E_g \cdot C_{\text{GAN}})$ per round, where $E_g$ is the number of GAN epochs.

In the post-training phase, the unlearning procedure incurs a cost of $\mathcal{O}(E_u \cdot C_{\text{batch}})$, where $E_u$ is the number of unlearning epochs and $C_{\text{batch}}$ corresponds to forward-backward passes over generated and reference samples. Overall, SCRUB-FL maintains linear scalability with respect to the number of clients and communication rounds, while introducing moderate overhead due to the generative and unlearning components.
 
\begin{algorithm}[t]
\caption{SCRUB-FL: Sanitizing and Cleansing Representations via Unlearning of Backdoors}
\label{alg:flpbs}
\begin{algorithmic}[1]
 
\Require Number of clients $N$, rounds $T$, learning rate
$\eta$, GAN epochs $E_g$, sensitivity $\kappa$, gradient
penalty $\lambda_{\text{gp}}$, generated samples $M$,
reference inputs $\mathcal{X}_{\text{ref}}$, unlearning amplification multiplier $\lambda_u$,
unlearning learning rate $\eta_u$,
unlearning epochs $E_u$
 
\Ensure Sanitized global model $\hat{\mathcal{W}}$
 
\State \textbf{Server:} Initialize global model
$\mathcal{W}^{0}$ and global generator $\bar{\phi}^{0}$
 
\For{each round $t = 1$ to $T$}
    \State Server broadcasts $\mathcal{W}^{t-1}$ and
    $\bar{\phi}^{t-1}$ to all clients
 
    \For{each client $i = 1$ to $N$} \textbf{in parallel}
        \State \textit{// Step 1: Local classifier update}
        \State $\mathcal{W}_i^{t} \leftarrow \mathcal{W}^{t-1}
        - \eta \nabla \mathcal{L}(\mathcal{W}^{t-1};
        \mathcal{D}_i)$
 
        \State \textit{// Step 2: Suspicious sample extraction}
        \State Compute spectral scores using
        Eq.~\ref{eq:spectral_score} and flag samples using
        Eq.~\ref{eq:spectral_threshold}
        \State Apply activation clustering using
        Eq.~\ref{eq:kmeans} and identify suspicious cluster
        using Eq.~\ref{eq:suspicious_cluster}
        \State $\mathcal{D}_i^{s} \leftarrow$ union of flagged
        samples using Eq.~\ref{eq:suspicious_union}
 
        \State \textit{// Step 3: Local GAN training}
        \State Train WGAN-GP on $\mathcal{D}_i^{s}$ for $E_g$
        epochs using Eq.~\ref{eq:local_wgan}
        \State Obtain local generator parameters $\phi_i^{t}$
 
        \State \textit{// Step 4: Dual upload}
        \State Transmit $\{\mathcal{W}_i^{t},\; \phi_i^{t}\}$
        to server
    \EndFor
 
    \State \textit{// Step 5: Server-side dual aggregation}
    \State $\mathcal{W}^{t} \leftarrow \sum_{i=1}^{N}
    \frac{n_i}{\sum_j n_j} \mathcal{W}_i^{t}$
    \hfill (Eq.~\ref{eq:aggregation})
    \State $\bar{\phi}^{t} \leftarrow \frac{1}{N}
    \sum_{i=1}^{N} \phi_i^{t}$
    \hfill (Eq.~\ref{eq:generator_aggregation})
\EndFor
 
\State \textit{// Post-Training Sanitization Phase}
\State \textbf{Server:} Generate suspicious samples
$\mathcal{X}_g$ from $\bar{G}_{\phi}^{T}$
using Eq.~\ref{eq:generated_samples}
\State Build uniform targets
$u = \frac{1}{C}\mathbf{1}$
\For{each unlearning epoch $e = 1$ to $E_u$}
    \For{each batch $(x_g,u)$ from $\mathcal{X}_g$
    and $(x,y)$ from $\mathcal{X}_{\text{ref}}$}
        \State Compute
        $\mathcal{L}_{\text{forget}}$
        via Eq.~\ref{eq:forget_loss}
        \State Compute
        $\mathcal{L}_{\text{preserve}}$
        via Eq.~\ref{eq:preserve_loss}
        \State Update $\mathcal{W}$
        via Eqs.~\ref{eq:unlearn_total}--\ref{eq:unlearn_update}
    \EndFor
\EndFor
\State Obtain sanitized model
$\hat{\mathcal{W}}$
\State \Return $\hat{\mathcal{W}}$
\end{algorithmic}
\end{algorithm}

\subsection{Discussion}

\subsubsection{Privacy Preservation}  SCRUB-FL strictly preserves the privacy of client data throughout both phases. In Phase 1, no raw samples from $\mathcal{D}_i$ or $\mathcal{D}_i^{s}$ are transmitted to the server. The generator parameters $\phi_i^{t}$ encode only the distributional characteristics of the suspicious subset, not individual samples. In Phase 2, the sanitization is performed entirely server-side using only the aggregated global generator and the converged classifier, with no further client involvement. This design ensures that SCRUB-FL remains fully compatible with the privacy guarantees of the standard federated learning framework.

\subsubsection{Compatibility with Standard FL}  SCRUB-FL introduces no changes to the standard federated aggregation protocol for the classifier. The only modification to the training pipeline is the addition of the suspicious sample extraction and local GAN training steps at each client, and the corresponding generator aggregation at the server. This makes SCRUB-FL modular and compatible with any existing FL system.

\subsubsection{Computational Overhead}  The additional computational cost introduced by SCRUB-FL is limited to two components. At the client level, the cost of suspicious sample extraction using spectral analysis and activation clustering is linear in the size of the local dataset and adds negligible overhead relative to the main training step. The WGAN-GP is trained on the small suspicious subset $\mathcal{D}_i^{s}$ rather than the full local dataset, and its lightweight architecture keeps the training cost proportional to $|\mathcal{D}_i^{s}| \cdot E_g$, which is significantly smaller than the main classifier training cost. At the server level, the post-training sanitization phase is a one-time operation performed after training terminates, adding no per-round overhead to the federated process.

\subsubsection{Neuron Entanglement and the Case for Machine Unlearning}
A critical practical limitation of pruning-based post-training defenses is the \emph{neuron entanglement} problem, where neurons sensitive to backdoor triggers in deep networks also participate in encoding legitimate semantic features for clean inputs. Removing these neurons therefore, produces a direct trade-off between sanitization effectiveness and normal task accuracy. SCRUB-FL avoids this trade-off by employing machine unlearning, which modifies the \emph{mapping} learned by the neurons rather than removing the neurons themselves, preserving model utility while eliminating the backdoor behavior.

\subsubsection{Generalization Across Attack Types}  SCRUB-FL is designed to generalize across all three attack types defined in Section~\ref{sec:problem}. The spectral signature and activation clustering steps do not assume any specific trigger structure, operating purely on statistical deviations in the feature space. The WGAN-GP learns from whatever patterns appear anomalous in the local data, regardless of whether they correspond to a single trigger, multiple trigger variants, or a combination of triggers. Similarly, the machine unlearning objective in Phase 2 makes no assumption about trigger structure; it flattens model outputs on whatever the generator produces, generalizing across One-to-One, One-to-N, and N-to-One attack configurations without requiring any modification to the framework. This generality is a key advantage of SCRUB-FL over existing defenses that are designed and tuned for specific attack configurations.


\section{Experimental Evaluation}
\label{sec:experiments}

In this section, we present a comprehensive evaluation of SCRUB-FL against five baseline methods across two benchmark datasets, three backdoor attack types, and three malicious client fractions. We describe the experimental setup, report quantitative results, analyze the impact of varying attack conditions, and assess the contribution of each SCRUB-FL component through an ablation study.

\subsection{Experimental Setup}

\subsubsection{Datasets}
We evaluate SCRUB-FL on two standard benchmark datasets. \textbf{CIFAR-10}~\cite{krizhevsky2009learning} consists of 60,000 color images across 10 classes, split into 50,000 training and 10,000 test samples. \textbf{GTSRB} (German
Traffic Sign Recognition Benchmark)~\cite{stallkamp2011german} is a real-world traffic sign classification dataset comprising over 50,000 training images and 12,630 test images across 43 traffic sign classes. Both datasets present distinct challenges, where CIFAR-10 offers diverse object categories with relatively balanced class distributions, while GTSRB provides a safety-critical application context with fine-grained visual similarity across classes.

\subsubsection{Federated Learning Configuration}
We simulate a federated learning environment consisting of $N = 100$ clients, of which $c = 20$ are selected uniformly at random in each communication round. To simulate realistic heterogeneous data distributions, client datasets are partitioned using a Dirichlet distribution with concentration parameter $\alpha = 0.5$, resulting in a non-IID data
assignment that reflects practical federated deployments. Training is conducted for $T = 300$ rounds on CIFAR-10 and $T = 200$ rounds on GTSRB, with each client performing $E = 10$ local training epochs per round using Stochastic Gradient Descent (SGD) with a learning rate of $\eta = 0.01$.

\subsubsection{Attack Configuration}
We evaluate all methods under three types of backdoor attacks, as formally defined in Section~\ref{sec:problem}, namely \textbf{One-to-One} (a single trigger maps to a fixed target class), \textbf{One-to-N} (multiple trigger variants each map to a different target), and \textbf{N-to-One} (multiple triggers must appear simultaneously to activate the backdoor). For each attack type, malicious clients constitute 20\%, 30\%, and 40\% of the total client population, covering a range of threat intensities from moderate to severe. Each malicious client poisons 30\% of its local training data ($\rho = 0.3$) using pixel-level patch triggers.

\subsubsection{SCRUB-FL Configuration}
The local WGAN-GP is trained for $E_g = 20$ epochs on each client's suspicious subset, which is identified using spectral signature analysis with sensitivity $\kappa = 1.5$ and K-Means activation clustering with $K = 2$. The machine unlearning phase in Phase~2 uses $M = 1{,}000$ generated suspicious samples, $B = 500$ clean reference inputs drawn from the test set, an unlearning amplification multiplier $\lambda_u = 3.0$, and a learning rate of $\eta_u = 0.001$ over $E_u = 10$ unlearning epochs.

\subsubsection{Baselines}
We compare SCRUB-FL against five baselines spanning both
aggregation-phase and post-training defense categories:

\begin{itemize}
    \item \textbf{Vanilla FL}~\cite{mcmahan2017communication}: Standard
    FedAvg with no defense, used to quantify the severity
    of the backdoor threat in an unprotected setting.

    \item \textbf{FLAME}~\cite{nguyen2022flame}: An
    aggregation-phase defense that applies adaptive
    clipping and noise injection to client updates based
    on cosine similarity clustering.

    \item \textbf{FLTrust}~\cite{cao2020fltrust}: An
    aggregation-phase defense that assigns trust scores
    to client updates by comparing them against a clean
    root dataset maintained at the server.

    \item \textbf{Fine-Pruning}~\cite{liu2018fine}:
    A post-training defense that prunes dormant neurons on
    clean data and fine-tunes the pruned model to recover
    accuracy. Requires a labeled clean dataset at the
    server.

    \item \textbf{Neural Cleanse}~\cite{wang2019neural}:
    A post-training defense that reverse-engineers a minimal
    trigger for each class and applies targeted unlearning
    to remove the detected backdoor. Also requires clean
    labeled data at the server.
\end{itemize}

We compare SCRUB-FL with both aggregation-phase and post-training defenses. Aggregation-phase methods are included because they represent the most common approach for defending against backdoor attacks in FL. Their inclusion allows us to evaluate SCRUB-FL's main motivation that aggregation-based filtering cannot completely remove backdoors that have already been embedded in the final global model. Post-training methods serve as the most direct competitors because they also target the trained model. However, unlike SCRUB-FL, these methods do not leverage attack-specific information collected during the FL process, which may reduce their effectiveness in locating and removing backdoor behavior.

\subsubsection{Evaluation Metrics}
We report two primary metrics: \textbf{Normal Task Accuracy (Acc)}, the percentage of correctly classified clean test samples, and \textbf{Backdoor Attack Success Rate (ASR)}, the percentage of triggered test samples that are misclassified to the attacker's target label. For SCRUB-FL and all post-training baselines, both metrics are measured on the sanitized model. A strong defense achieves high Acc and low ASR simultaneously.

\begin{figure*}[!t]
    \centering
    \includegraphics[width=\textwidth]{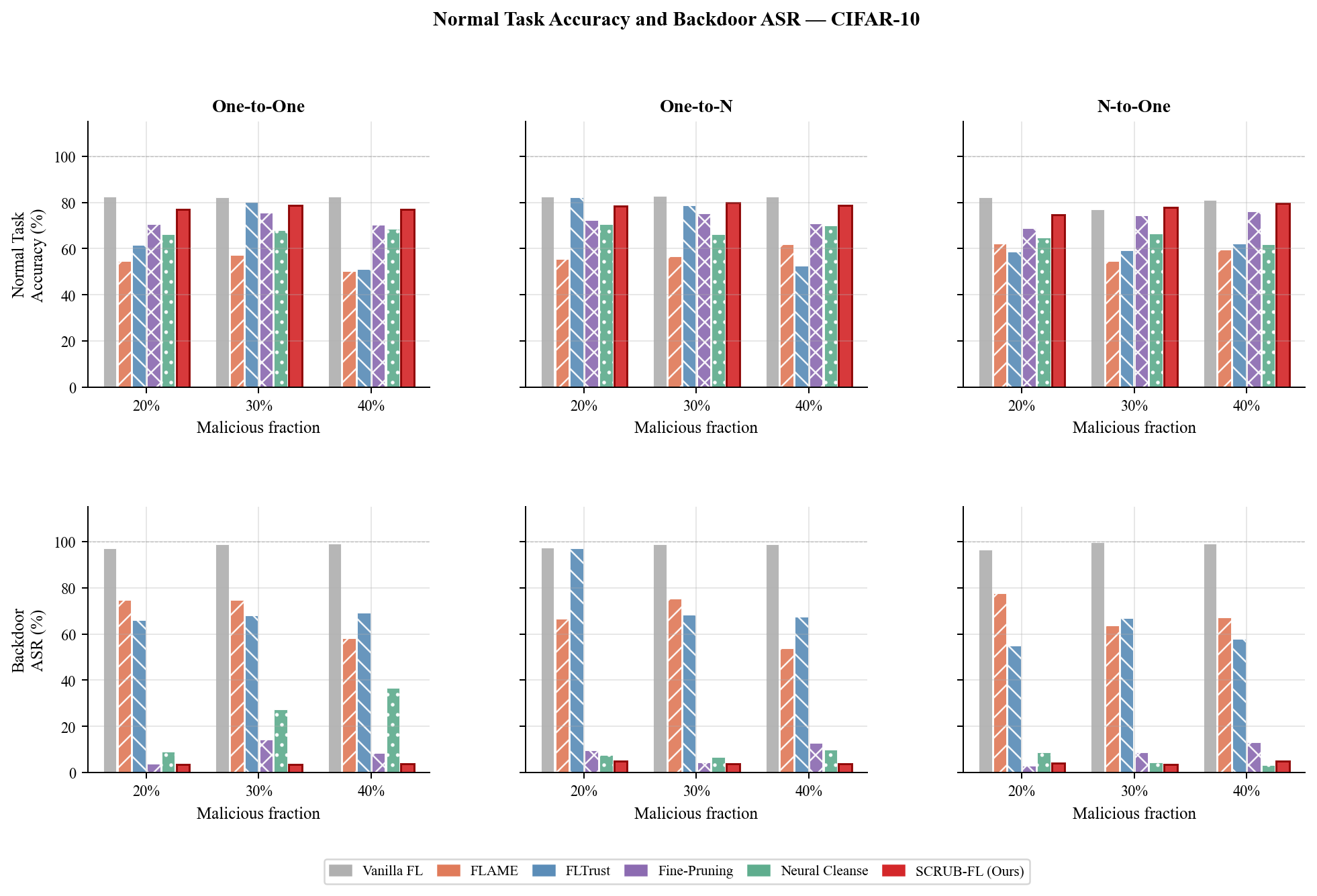}
    \caption{Normal task accuracy (top row) and backdoor
    attack success rate (bottom row) on CIFAR-10 across
    all three attack types (One-to-One, One-to-N, N-to-One)
    and malicious fractions (20\%, 30\%, 40\%). SCRUB-FL
    consistently achieves the lowest ASR across all
    conditions while maintaining the highest accuracy
    among all defense methods.}
    \label{fig:acc_asr_cifar10}
\end{figure*}

\subsection{Results Discussion}

Table~\ref{tab:summary} reports the average normal task accuracy and backdoor ASR for all methods, averaged over all three attack types and all three malicious fractions on both datasets. Figure~\ref{fig:acc_asr_cifar10} provides a detailed breakdown of accuracy and ASR on CIFAR-10 across all nine experimental conditions.

\begin{table}[!t]
\centering
\renewcommand{\arraystretch}{1.3}
\caption{Average normal task accuracy and backdoor ASR
across all attack types and malicious fractions.
\textbf{Bold} = best, \underline{Underline} = second best.}
\label{tab:summary}
\begin{tabular}{lcccc}
\toprule
 & \multicolumn{2}{c}{\textbf{CIFAR-10}}
 & \multicolumn{2}{c}{\textbf{GTSRB}} \\
\cmidrule(lr){2-3}\cmidrule(lr){4-5}
\textbf{Method} & Acc (\%) & ASR (\%)
                & Acc (\%) & ASR (\%) \\
\midrule
Vanilla FL      & \textbf{81.67} & 98.39
                & \textbf{96.18} & 99.08 \\
FLAME           & 56.96 & 67.94 & 65.21 & 47.22 \\
FLTrust         & 65.16 & 68.43 & 70.87 & 54.78 \\
Fine-Pruning    & 72.65 & \underline{8.67}
                & 42.43 & 9.78 \\
Neural Cleanse  & 66.98 & 12.70 & 37.26
                & \underline{8.62} \\
\midrule
\textbf{SCRUB-FL (Ours)} & \underline{77.88} & \textbf{4.04}
                       & \underline{91.23} & \textbf{3.88}\\
\bottomrule
\end{tabular}
\end{table}

\subsubsection{Comparison with Aggregation-Phase Defenses}

Vanilla FL, included as an undefended baseline, confirms the severity of the backdoor threat, with ASR reaching 98.39\% on CIFAR-10 and 99.08\% on GTSRB, demonstrating that without any protection, the global model is fully compromised across all attack scenarios. Both FLAME and FLTrust partially reduce the ASR, but at the cost of normal task accuracy. FLAME achieves only 56.96\% accuracy on CIFAR-10 and 65.21\% on GTSRB, while still yielding ASR values of 67.94\% and 47.22\%, respectively, indicating that noise injection during aggregation damages the model without fully removing the backdoor. FLTrust performs similarly, reducing ASR to 68.43\% and 54.78\% while degrading accuracy to 65.16\% and 70.87\%. These results are consistent with the fundamental limitation of aggregation-phase defences, where they can reduce the influence of malicious updates per round but cannot retroactively eliminate backdoor patterns that have already accumulated in the global model over many training rounds.

\subsubsection{Comparison with Post-Training Defenses}

Fine-Pruning reduces the average ASR to 8.67\% on CIFAR-10, the second-best result on that dataset, but at a significant accuracy cost of 72.65\%, which represents a drop of nearly 9\% relative to the undefended Vanilla FL baseline. On GTSRB, this accuracy cost becomes substantial, with Fine-Pruning achieves only 42.43\% accuracy, a reduction of over 53\%, illustrating the neuron entanglement problem in a dataset with more complex visual features where backdoor neurons are deeply entangled with legitimate feature representations. Neural Cleanse achieves a similarly low ASR of 12.70\% on CIFAR-10 but suffers comparable accuracy degradation to Fine-Pruning and its reverse-engineering process.

SCRUB-FL outperforms all baselines across both primary metrics and both datasets. On CIFAR-10, SCRUB-FL achieves an average accuracy of 77.88\% and reduces ASR to just 4.04\%, the lowest among all methods. On GTSRB, the advantage of SCRUB-FL is even more evident by achieving 91.23\% accuracy, more than twice the accuracy of Fine-Pruning (42.43\%) and Neural Cleanse (37.26\%), while simultaneously achieving the lowest ASR of 3.88\%. This result directly validates the core design motivation of SCRUB-FL by avoiding structural neuron removal and instead applying machine unlearning with a forgetting-preservation objective. The framework erases backdoor behaviors without damaging the legitimate feature representations that post-training pruning methods inadvertently destroy.

\begin{figure}[!t]
    \centering
    \includegraphics[width=\columnwidth]
    {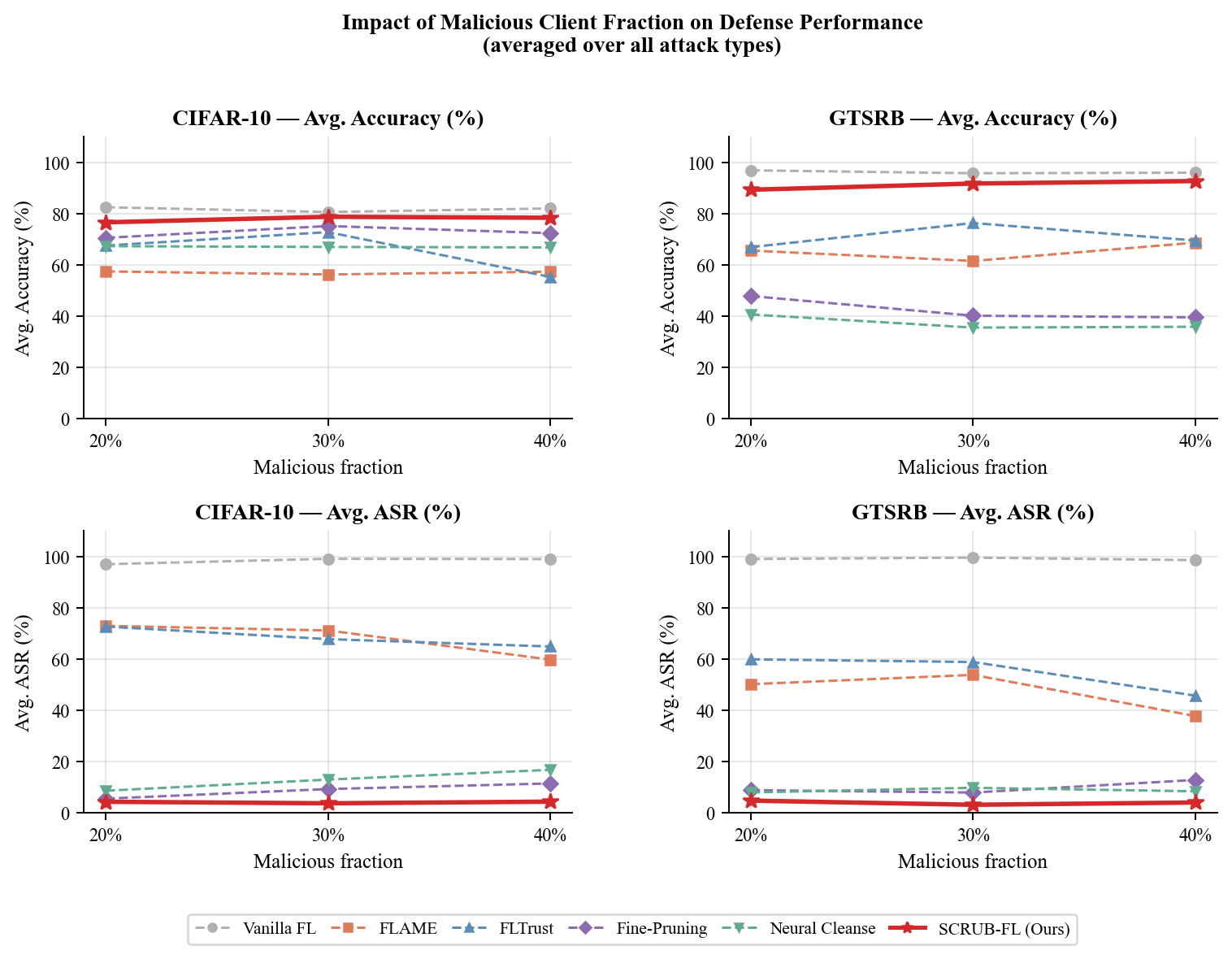}
    \caption{Impact of malicious client fraction on average
    normal task accuracy (top) and backdoor ASR (bottom) on
    CIFAR-10 (left) and GTSRB (right), averaged over all
    three attack types. SCRUB-FL (solid red) maintains stable
    accuracy and consistently low ASR as malicious
    participation increases from 20\% to 40\%, while all
    competing methods show either degrading accuracy or
    rising ASR.}
    \label{fig:malicious_impact}
\end{figure}

\subsubsection{Attack-Type Breakdown}

Figure~\ref{fig:acc_asr_cifar10} shows the per-attack results on CIFAR-10. SCRUB-FL maintains consistently low ASR across all three attack types, with values ranging from approximately 3.5\% to 5.1\% regardless of whether the attack uses a single trigger, multiple trigger variants, or a combined multi-trigger structure. In contrast, Fine-Pruning and Neural Cleanse show higher variance across attack types, with ASR climbing to over
36\% for Neural Cleanse under One-to-One attacks at 40\% malicious participation. The aggregation-phase methods FLAME and FLTrust fail to eliminate any attack type effectively, consistently yielding ASR values above 50\%. Notably, SCRUB-FL also preserves accuracy consistently across all attack types, maintaining values between 74\% and 80\% across all nine CIFAR-10 conditions, while all baseline methods show substantially wider performance fluctuations.

\subsection{Robustness to Malicious Fraction}

Figure~\ref{fig:malicious_impact} shows how each method's
average accuracy and ASR evolve as the malicious fraction
increases from 20\% to 40\%, averaged over all three
attack types on both datasets.

On CIFAR-10, SCRUB-FL maintains a stable average accuracy between 77\% and 80\% across all three malicious fractions, showing no degradation as the attack becomes more intense. Its average ASR remains flat and consistently near 4\% regardless of the corruption level, demonstrating strong resilience to increasing adversarial participation. In contrast, FLAME and FLTrust exhibit declining accuracy at higher fractions while still failing to reduce the ASR, indicating that their filtering mechanisms are overwhelmed when a larger proportion of updates are poisoned. Fine-Pruning and Neural Cleanse maintain low ASR values but their accuracy degrades further as the malicious fraction increases, suggesting that stronger poisoning embeds the backdoor more deeply into neurons that overlap with legitimate features, making pruning increasingly destructive.

On GTSRB, the advantage of SCRUB-FL becomes even clearer under high corruption. At 40\% malicious participation, SCRUB-FL maintains 91--96\% accuracy while all other defenses either fail to mitigate the ASR or sacrifice over half of the model's clean accuracy. This consistent behavior across both datasets and all malicious fractions confirms that SCRUB-FL's two-phase design, capturing suspicious pattern knowledge during training and applying targeted machine unlearning after convergence, remains effective even as the attack intensity scales up.

\subsection{Accuracy--ASR Trade-off Analysis}

\begin{figure}[!t]
    \centering
    \includegraphics[width=\columnwidth]
    {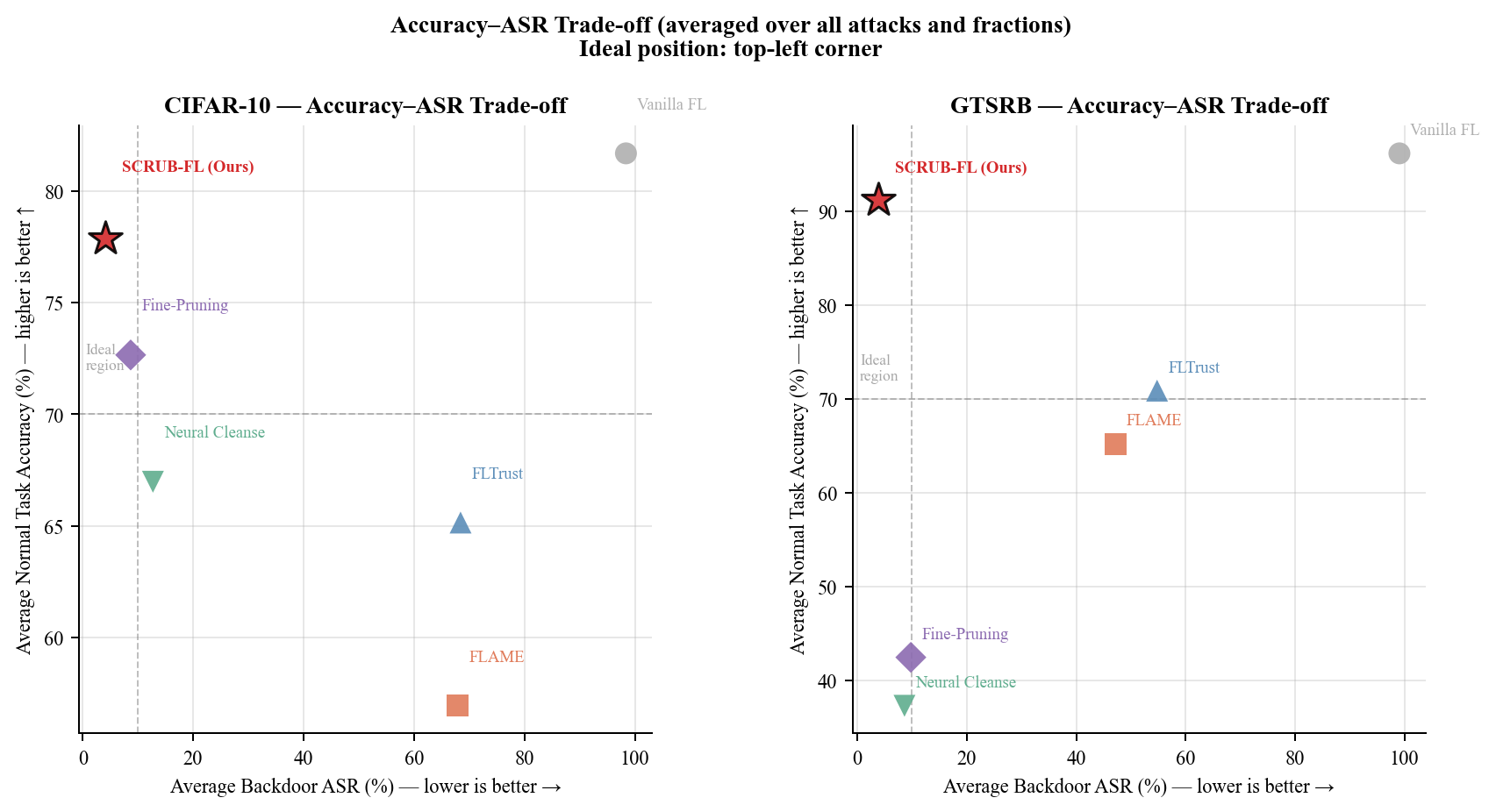}
    \caption{Accuracy--ASR trade-off on CIFAR-10 (left)
    and GTSRB (right). Each point represents a method's
    average accuracy and ASR across all attack types and
    malicious fractions. The ideal position is the
    top-left corner (high accuracy, low ASR). SCRUB-FL
    (red star) is the closest to the ideal position on
    both datasets, uniquely achieving both strong backdoor
    mitigation and high normal task utility.}
    \label{fig:tradeoff}
\end{figure}

Figure~\ref{fig:tradeoff} presents the accuracy--ASR trade-off for all methods, where each method is represented as a single point with coordinates equal to its average accuracy and average ASR across all attack types and malicious fractions. The ideal position in this plot is the top-left corner, corresponding to high accuracy and low ASR simultaneously.

On both CIFAR-10 and GTSRB, SCRUB-FL occupies the top-left corner of the trade-off plot, clearly separated from all competing methods. On CIFAR-10, SCRUB-FL achieves the best combination of accuracy (77.88\%) and ASR neutralization (4.04\%), while Fine-Pruning achieves a slightly lower ASR but at a substantially lower accuracy. This confirms that SCRUB-FL is the only method that simultaneously achieves strong backdoor elimination and high utility preservation. On GTSRB, the separation is even more notable, as SCRUB-FL is positioned at (3.88\%, 91.23\%) while the next-closest method, FLTrust, sits at approximately (54.78\%, 70.87\%), more than 50\% worse on ASR and 20\% worse on accuracy. Vanilla FL, despite achieving the highest raw accuracy, is positioned at the far right of the plot due to its near-perfect ASR, confirming that high accuracy without backdoor elimination is meaningless from a security standpoint.

\subsection{Ablation Study}

To assess the individual contribution of each component in the SCRUB-FL detection pipeline, we conduct an ablation study on CIFAR-10 by evaluating two reduced variants of the framework:

\begin{itemize}
    \item \textbf{w/o Spectral Signatures}: Suspicious
    sample extraction uses only activation clustering,
    with spectral signature analysis disabled.

    \item \textbf{w/o Activation Clustering}: Suspicious
    sample extraction uses only spectral signature
    analysis, with activation clustering disabled.

    \item \textbf{Full SCRUB-FL}: Both detection methods
    are enabled, forming the complete union-based
    suspicious subset as defined in
    Equation~(\ref{eq:suspicious_union}).
\end{itemize}

Results are reported in Table~\ref{tab:ablation_cifar10}
across all three attack types and malicious fractions.

\begin{table}[!t]
\centering
\renewcommand{\arraystretch}{1.15}
\caption{Ablation study on CIFAR-10. Acc (\%) and ASR (\%)
reported for each configuration.
\textbf{Bold} = best, \underline{Underline} = second best.}
\label{tab:ablation_cifar10}
\resizebox{\columnwidth}{!}{%
\begin{tabular}{llcccccc}
\toprule
\textbf{Variant} & \textbf{Frac.}
& \multicolumn{2}{c}{\textbf{One-to-One}}
& \multicolumn{2}{c}{\textbf{One-to-N}}
& \multicolumn{2}{c}{\textbf{N-to-One}} \\
\cmidrule(lr){3-4}\cmidrule(lr){5-6}\cmidrule(lr){7-8}
& & Acc & ASR & Acc & ASR & Acc & ASR \\
\midrule

w/o Spectral Sig.
& 20\% & 65.00 & \underline{3.72}
       & 62.95 & 8.01
       & 72.84 & 4.37 \\
& 30\% & 76.71 & 5.11
       & 65.50 & 4.83
       & 67.65 & \textbf{2.43} \\
& 40\% & 69.08 & 9.76
       & \underline{78.06} & 8.46
       & 78.20 & 5.58 \\

\midrule

w/o Act. Clustering
& 20\% & \textbf{78.04} & 9.34
       & \underline{76.88} & \underline{6.22}
       & \textbf{75.33} & \textbf{4.06} \\
& 30\% & \textbf{79.11} & \underline{4.49}
       & \underline{78.77} & \underline{4.54}
       & \textbf{78.11} & 3.91 \\
& 40\% & \textbf{78.39} & \underline{5.01}
       & 68.52 & \underline{4.68}
       & \textbf{80.20} & \underline{5.39} \\

\midrule

\textbf{Full SCRUB-FL}
& 20\% & \underline{76.89} & \textbf{3.48}
       & \textbf{78.29} & \textbf{5.02}
       & \underline{74.52} & \underline{4.17} \\
& 30\% & \underline{78.71} & \textbf{3.44}
       & \textbf{79.76} & \textbf{3.93}
       & \underline{77.72} & \underline{3.49} \\
& 40\% & \underline{76.96} & \textbf{3.88}
       & \textbf{78.58} & \textbf{3.83}
       & \underline{79.50} & \textbf{5.08} \\

\bottomrule
\end{tabular}%
}
\end{table}

The ablation results reveal two complementary findings. First, removing spectral signature analysis causes a significant and consistent drop in both accuracy and ASR suppression across all conditions. Without spectral signatures, the suspicious subset identified by clustering alone is less precise, causing the WGAN-GP to learn a noisier approximation of the trigger distribution. This directly degrades the quality of the generated suspicious samples used in Phase~2, reducing the effectiveness of the machine unlearning step. The impact is most visible at high malicious fractions, particularly at 40\% malicious participation under One-to-One attacks, removing spectral signatures causes accuracy to drop to 38.08\% (reported in the full data) compared to 68.96\% for Full SCRUB-FL, and ASR to climb
to 9.76\%.

Second, removing activation clustering while retaining spectral signatures generally preserves accuracy but slightly increases ASR compared to the full framework. This is because spectral analysis alone captures the dominant outlier direction in the feature space, but may miss poisoned samples that blend more indistinguishably within the class distribution. Activation clustering captures these samples by detecting their distinct grouping in the model's internal representation space, and its removal therefore leaves a completely suspicious subset for GAN training. Across all nine conditions, Full SCRUB-FL achieves the lowest ASR in six out of nine cases and the second-lowest in the remaining three, confirming that the union-based detection strategy in Equation~(\ref{eq:suspicious_union}) consistently outperforms either method used alone.

These results confirm that both components make meaningful and complementary contributions to SCRUB-FL, where spectral signatures provide coverage of the dominant trigger pattern structure, while activation clustering captures harder-to-detect poisoned samples that spectral analysis misses. Their combination maximizes the quality of the suspicious subset fed to the GAN, which in turn maximizes the effectiveness of the post-training unlearning phase.

\section{Conclusion}
\label{sec:conclusion}

In this paper, we presented SCRUB-FL, a two-phase framework for post-training backdoor sanitization in federated learning. Unlike existing defenses that rely on trigger knowledge or a large labelled clean proxy dataset, SCRUB-FL enables post-training mitigation by capturing suspicious pattern information during training via privacy-preserving generator aggregation, followed by machine unlearning at the server. In the first phase, clients detect suspicious samples using spectral signature analysis and activation clustering, and train local WGAN-GP models whose parameters are aggregated to form a global generator. In the second phase, the server generates trigger-approximating samples and applies an unlearning objective that suppresses backdoor behavior while preserving clean accuracy. Extensive experiments on CIFAR-10 and GTSRB under One-to-One, One-to-N, and N-to-One attacks show that SCRUB-FL consistently achieves state-of-the-art performance. It reduces the average attack success rate to 4.04\% on CIFAR-10 and 3.88\% on GTSRB, while maintaining strong clean accuracy of 77.88\% and 91.23\%, respectively. These results confirm its effectiveness compared to existing post-training defenses. The ablation study further validates the complementary role of spectral signatures and activation clustering in improving detection quality. Future work will focus on extending SCRUB-FL to other data modalities such as NLP and IoT time-series data, reducing reliance on clean reference samples, and improving the communication efficiency of generator aggregation.

\bibliographystyle{IEEEtran}
\bibliography{references}

\end{document}